\documentclass[letterpaper, 10 pt, conference]{ieeeconf} 
\IEEEoverridecommandlockouts                              
\usepackage[usenames,dvipsnames,table,xcdraw]{xcolor}
\usepackage[utf8]{inputenc}
\usepackage{graphicx}
\usepackage{multicol}
\usepackage{multirow}
\usepackage{array}
\usepackage{footmisc}
\usepackage{amssymb}
\usepackage{amsmath}
\usepackage{amsfonts}
\usepackage{hyperref}
\usepackage{algorithm}
\usepackage{tikz}
\usepackage{float}
\usetikzlibrary{math,shapes,calc,decorations.text}
\usepackage[caption=false, font=footnotesize]{subfig}
\usepackage{listings}
\usepackage{xcolor}
\usepackage{booktabs} %
\usepackage{algpseudocode}
\usepackage[backend=biber,
            url=false,
            isbn=false,
            doi=false,
            backref=false,
            style=ieee,
            natbib=true,
            mincitenames=1,
            maxcitenames=1,
            citestyle=numeric-verb,
            sorting=none,%
            block=none]{biblatex}
\renewcommand{\bibfont}{\small}
\bibliography{./references.bib}

\DeclareMathOperator*{\argmin}{\arg\!\min}

\usepackage{amsmath}
\DeclareMathOperator*{\argmax}{arg\,max}

\title{\LARGE \bf
Self-Supervised Learning for Interactive Perception \\ of Surgical Thread for Autonomous Suture Tail-Shortening
}

\author{Vincent Schorp$^{1,2}$, Will Panitch$^{1}$, Kaushik Shivakumar$^{1}$, Vainavi Viswanath$^{1}$, Justin Kerr$^{1}$ \\Yahav Avigal$^{1}$,  Danyal M Fer$^{3}$, Lionel Ott$^{2}$, Ken Goldberg$^{1}$
\thanks{The AUTOLab at the University of California, Berkeley (\url{automation.berkeley.edu}), 
{\tt\small \{vschorp, goldberg\} @berkeley.edu}}
\thanks{$^{1}$ AUTOLAB at University of California, Berkeley} 
\thanks{$^{2}$ Autonomous Systems Lab, ETH Zurich}
\thanks{$^{3}$ MD, Department of Surgery, University of California San Francisco East Bay}
}

\begin{document}

\maketitle
\thispagestyle{empty}
\pagestyle{empty}

\begin{abstract}
Accurate 3D sensing of suturing thread is a challenging  problem in automated surgical suturing because of the high state-space complexity, thinness and deformability of the thread, and possibility of occlusion by the grippers and tissue. In this work we present a method for tracking surgical thread in 3D which is robust to occlusions and complex thread configurations, and apply it to autonomously perform the surgical suture ``tail-shortening'' task: pulling thread through tissue until a desired ``tail'' length remains exposed. The method utilizes a learned 2D surgical thread detection network to segment suturing thread in RGB images. It then identifies the thread path in 2D and reconstructs the thread in 3D as a NURBS spline by triangulating the detections from two stereo cameras. Once a 3D thread model is initialized, the method tracks the thread across subsequent frames. Experiments suggest the method achieves a 1.33 pixel average reprojection error on challenging single-frame 3D thread reconstructions, and an 0.84 pixel average reprojection error on two tracking sequences. On the tail-shortening task, it accomplishes a 90\% success rate across 20 trials. Supplemental materials are available at: \url{https://sites.google.com/berkeley.edu/autolab-surgical-thread/} 

\end{abstract}

\section{Introduction}

Many steps in suturing, such as tail-shortening (where a thread is pulled through a suture to a desired length) and knot tying, require accurate thread tracking, which is particularly challenging due to the thin and flexible nature of suturing thread, as well as its propensity for self-intersections and partial occlusions. 

\begin{figure}
    \centering
    \vspace{0.08in}
    \includegraphics[width=1.0\linewidth]{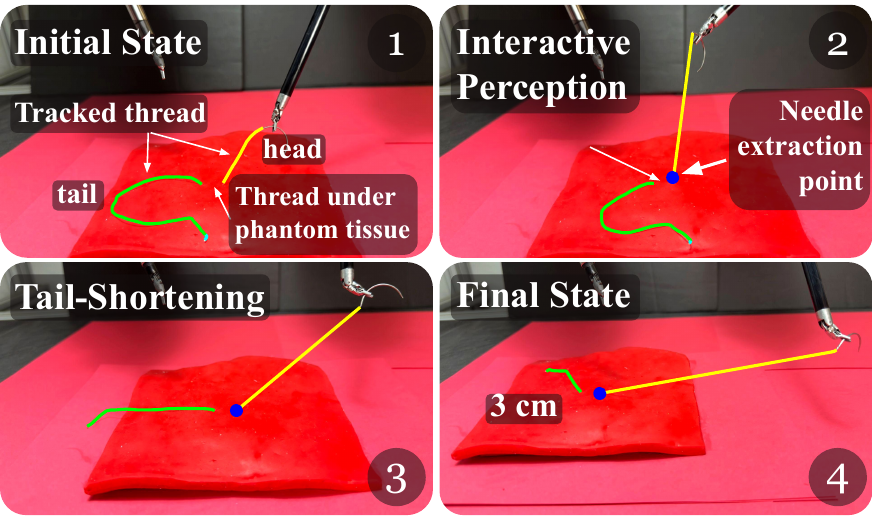}
    \vspace{-0.25in}
    \caption{\textbf{Surgical Suture Tail-shortening with 3D Thread Tracking.} We present a thread tracking method which we use for automating surgical suture ``tail-shortening'', i.e. pulling thread through tissue until a desired length of thread remains exposed. Initially, the robot grasps the needle close to the wound. It uses interactive perception to determine which portions of the reconstructed thread are on the needle side (head) vs slack side (tail). The system accomplishes tail-shortening by visually servoing the needle driver until a desired tail length remains.}
    \label{fig:splash}
     \vspace{-0.25in}
\end{figure}

In this paper, we propose a novel interactive perception system for tracking suture thread in 3D, which we apply to track thread in the autonomous tail-shortening task described in Figure~\ref{fig:splash}, requiring precise tracking to avoid pulling the thread too far out of the suture.

The learned 2D suturing thread detection model is trained using the Labels from UltraViolet (LUV) method~\parencite{luv} for self-supervised data collection, which has previously been shown to be effective for detecting cables and surgical needles. We extend it to surgical threads and combine the detection network with a 3D tracking method for temporal stability. Modeling the thread in 3D is a non-trivial task due to its complex shape and unclear endpoints. To address this issue, we model the thread as a 3D Non-Uniform Rational B-Spline (NURBS)~\parencite{thread_reconstruction_cavusoglu} based on stereo images of the scene. We adapt the thread model across frames by optimizing spline control points to minimize the error between the current detections and the reprojection of the 3D spline into the images.

 Using NURBS optimization alone can break down in complex thread configurations because of false-positive detections or self-intersections. To address this, we develop an analytic 2D tracing approach based on prior work for cable untangling~\parencite{sgtm2}, which is used as a prior to prevent the NURBS optimization from collapsing in the presence of distracting false detections or challenging thread configurations.

\begin{figure*}[!ht]
    \vspace{0.03in}
    \centering
    \includegraphics[width=0.999\linewidth]{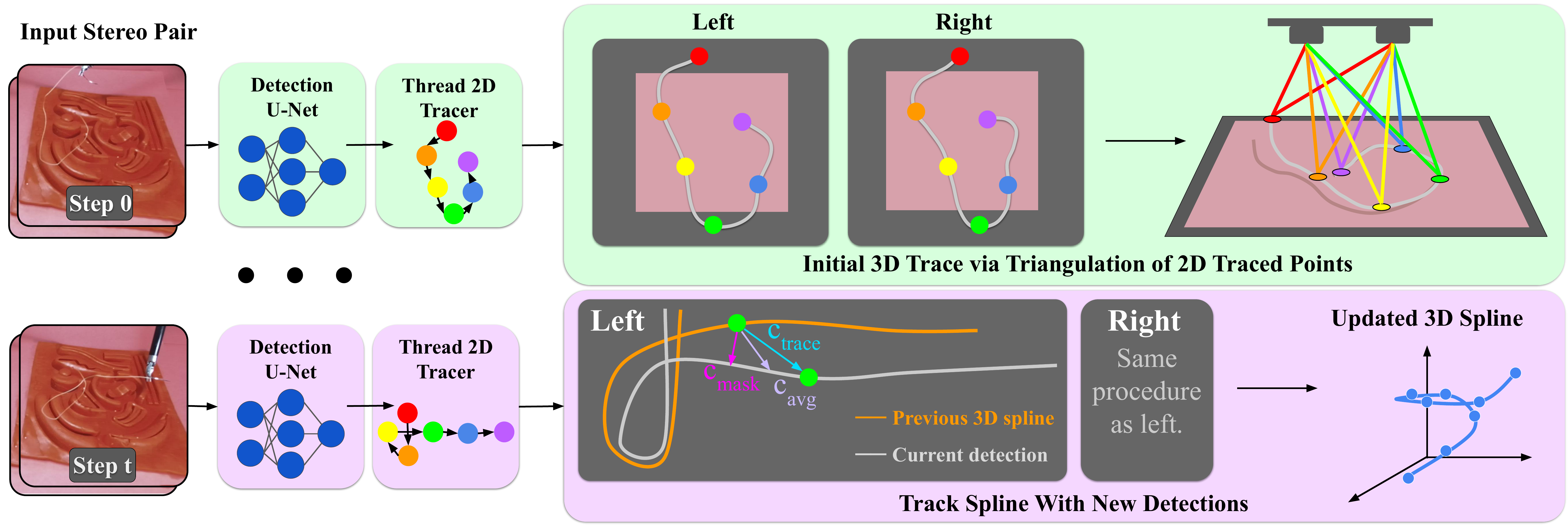}
    \vspace*{-0.25in}
    \caption{\textbf{Overview of the first 4 modules: 2D Surgical Thread Detection, 2D Tracing, 3D Tracing, and 3D Tracking}. \textit{Left:} For every stereo pair of images, we predict thread segmentation masks, then run a 2D tracer to compute the sequence of pixels along the thread. \textit{Top right:} To initialize the 3D spline of the thread, we match points meeting both stereo image and tracer topology constraints and triangulate their positions in 3D. We initialize the 3D trace by fitting a 3D spline to these points. \textit{Bottom right:} To update the 3D spline with new frames, we compute correction vectors in 2D as an average of vectors which push the projected 3D spline onto the new detection and push each projected 3D point to its corresponding point on the 2D trace. We then triangulate the correction vectors across both images and apply them to the 3D spline to perform an update.}
    \label{fig:methods}
    \vspace*{-0.25in}
\end{figure*}

Experiments conducted on a physical Intuitive Surgical da Vinci Research Kit (dVRK) RSA demonstrate that the system utilizing our thread tracker achieves 18/20 successful trials on the tail-shortening task (shown in Figure~\ref{fig:splash}).

This paper makes the following contributions:
\begin{enumerate}
    \item A 3D surgical thread tracking algorithm, described in Figure~\ref{fig:methods}, that combines a learned thread detection module trained on data collected in a self-supervised fashion with a NURBS spline optimization. 
    \item An interactive perception approach to suture thread tail-shortening which utilizes the thread tracker with visibility-maximizing manipulation to estimate the length of remaining thread tail.
    \item Data from experiments evaluating the perception components individually and applied to suture tail-shortening, achieving a 90\% success rate.
\end{enumerate}

\section{Related Work}

\subsection{Surgical Thread Detection}
Detecting surgical thread from an RGB image has been previously explored in a number of different settings. Early approaches relying on analytic curvilinear detectors~\parencite{thread_reconstruction_hager_1, thread_reconstruction_cavusoglu} work well when the thread is isolated and clearly visible, but fail in realistic scenes with shadows and occlusions. Similarly, \textcite{thread_reconstruction_yip} assume that thread detections can be obtained from color segmentation; however, this may fail due to light glare, sensor noise, materials covering the thread (e.g., blood), and varying lighting conditions.
Learning-based approaches generalize better to different backgrounds and lighting conditions, but require manual collection of large datasets. \textcite{thread_reconstruction_liu} train a U-Net~\parencite{ronneberger2015u} using semi-supervised learning leveraging hand-labeled images for supervision which are time consuming to obtain. 
We use a self-supervised data collection method that extracts labels autonomously using UV light~\parencite{luv}, allowing the system to collect 10 labeled images per minute.

\subsection{Surgical Thread Reconstruction}
\textcite{thread_reconstruction_liu}  propose using a 3D graph to represent the triangulated 3D candidate thread points. The method then computes a minimum energy path through the graph and uses it as the 3D model of the thread. 
\textcite{thread_reconstruction_yip} propose using a minimum variation spline to represent the suture. This results in a smooth reconstruction with less tight curvature and yields a confidence value along the spline model which is useful to chose a grasp point along the thread.
Both methods mentioned above fully reconstruct the model on each frame, ignoring prior frames, making them more susceptible to one-off missing or false detections.
\textcite{thread_reconstruction_hager_1}
assume the 3D spline has been initialized in advance, and focus on tracking the spline across frames. However, this work assumes that the length of the thread is constant, which limits its applicability to certain applications like tail shortening or knot tying.
\textcite{thread_reconstruction_cavusoglu} propose an approach to jointly trace and reconstruct a 3D spline from stereo images as well as a tracking method using pixel-space error minimization.
However, their approach assumes a known initial tracing point, manually defined using a space mouse. 
In contrast, we leverage tracing of 2D splines to address missing or occluded parts of the thread and use an approach which does not rely on a user-defined seed point.
Furthermore, our method tracks the spline across frames, increasing its robustness to noisy detections.

\subsection{Interactive Perception}
\textcite{goldberg1984active} investigate how a robot can use active perception to recognize the shape of an object by moving a touch sensor to trace its contours. \textcite{bajcsy1988active} defines \textit{active perception} as the search for models and control strategies for perception which can vary depending on the sensor and the task goal, such as adjusting camera parameters~\parencite{bajcsy2018revisiting} or moving a tactile sensor in response to haptic input~\parencite{goldberg1984active}. 

Similarly, \textit{interactive perception}, as explored by \textcite{bohg2017interactive}, utilizes robot interactions to enhance perception. Interactive perception has been used in robotic manipulation to extract kinematic and dynamic models from physical interactions with the environment~\parencite{martin2022coupled} and to improve the understanding of a scene in the presence of occlusions and perception uncertainty~\parencite{bohg2017interactive, danielczuk2019mechanical, novkovic2020object}. \textcite{murali2022active} leverage feedback from visual and tactile sensors to estimate the pose of partially occluded objects in cluttered environments. \textcite{danielczuk2019mechanical} propose the mechanical search problem, where a robot retrieves an occluded target object from a cluttered bin through a series of targeted parallel jaw grasps, suction grasps, and pushes. \textcite{novkovic2020object} use a robot to move a camera and interact with the environment in order to find a hidden target cube in a pile of cubes, while \textcite{shivakumar2022sgtm} use interactive perception to reduce perception uncertainty when untangling long cables. 

In this work we propose an interactive perception-based approach to surgical suture tail-shortening.  The robot tensions the thread to create a sharp angle between the taut thread on the extraction side of the suture and the slack thread on the insertion side.  This forces the thread into a linear configuration to facilitate perception. The robot then pulls the thread through the suture until the desired length of slack thread is detected at the tail.

\section{Problem Statement}
Using stereo RGB images, we want to accurately track the state of a surgical thread and use these state estimates to automate the task of surgical tail-shortening.

\subsection{Workspace and Assumptions}
\label{sec:workspace_setup}
We define the workspace using a cartesian $(x,y,z)$ coordinate system. The workspace consists of a bimanual dVRK robot \parencite{Kazanzidesf2014}; a Simulab TSP-10 human organ phantom\footnote{https://simulab.com/collections/suturing-skills-training/products/tissue-suture-pad}); and a fixed ZEDm RGB stereo camera, which outputs images at 1280x720 pixel resolution.
The camera is angled at the robot and phantom such that the whole reachable workspace of the robot is captured in the field of view.
We work with undyed (beige) $\text{Polysorb}^{\text{TM}}$ surgical suture thread from Covidien. The sutures are of variable length between 10 and 40 cm, with 2-0 USP Size (0.35-0.399 mm in diameter) and are attached to a GS-21 needle or similar. The length, diameter and needle size of the suture are unknown to the algorithm. 

We make the following assumptions:
\begin{enumerate}
    \item The robot-to-camera transform is known.
    \item  During test time, the thread and phantom configurations lie within the training data distribution. However, their pose does not necessarily correspond exactly to any pose seen in training.
    
\end{enumerate}

\section{Methods}

We decompose the problem of thread modeling and autonomous robot suture tail-shortening into five modules:
\begin{enumerate}
    \item \emph{Learned 2D Surgical Thread Detection}: uses a convolutional neural network to segment the surgical thread in a physical mockup of a surgical environment. This module takes as input an RGB image of dimension 1280 x 720 and returns a pixel-wise probability mask of the thread's location.
    \item \emph{2D Surgical Thread Tracing}: given the detection probability masks with potential gaps in the detections and false positives, identifies the sequence of image points along the thread.
    \item \emph{3D Surgical Thread Tracing}: computes a 3D representation of the suture thread based on the traced thread. This algorithm takes the traces from 2 rectified stereo images as input and returns a 3D NURBS spline.
    \item\emph{3D Surgical Thread Tracking}: adapts the 3D spline model to the current view of the scene. This module takes the traces from the current pair of rectified stereo images as well as the previous 3D spline as input and outputs an updated 3D spline.
    \item\emph{Surgical Suture Tail-Shortening}: This module performs the surgical tail-shortening task using an interactive perception approach which leverages the 3D spline model computed by the previous modules.
\end{enumerate}
An overview of how these modules are combined is shown in Figure~\ref{fig:methods}.

\subsection{Module 1: Learned 2D Surgical Thread Detection}
\label{method:detection}

We train a neural network to segment surgical thread from scenes using the self-supervised training approach proposed in \textcite{luv}. To collect labels automatically, we paint the surgical thread with a UV-florescent paint. This paint is invisible under visible light but shines when illuminated with UV light. For each scene, the robot arms are moved to a new random position in the workspace, changing the thread configuration and RGB stereo images are recorded under both visible and UV light. 

The label masks are extracted from the UV images using color segmentation. Train, validation, and test sets are split from disjoint subsets of scenes to ensure no cross-contamination of the sets from the same state. Using this self-supervised data collection technique, we are able to acquire 10 labeled images per minute (visible light stereo images with corresponding labels extracted under UV light). The image size of 1280 x 720 pixels is chosen to maximize the tradeoff between resolution (which aids in segmenting the thin thread) and inference speed (2.5 FPS on our test computer using an NVIDIA 2080 GPU).

\begin{figure}[!t]
\centering
\vspace{0.08in}
\includegraphics[width=1.0\linewidth]{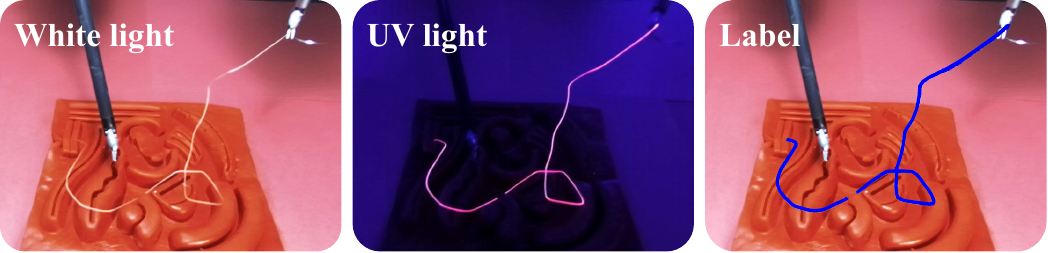}
\vspace*{-0.25in}
\caption{\textbf{2D Surgical Thread Detection Data Collection.} The left image shows the dVRK gripper holding a suture thread under white light. The middle image depicts the same scene under UV light. The thread painted with UV fluorescent color lights up and can be segmented via color thresholding. The right image displays the extracted label used for training.}
\label{fig:data_collection}
\vspace*{-0.25in}
\end{figure}

We train a U-Net \parencite{unet} to detect surgical thread from a single RGB image. We train our model on 1320 images of size 1280 x 720 for 400 epochs. We specifically choose not to upweight false negatives, as would be expected from the ratio of background pixels to thread pixels, as this yields predictions that are biased towards precision over recall. This is desirable because the 2D tracer is able to bridge missing thread detections but can get confused by false positives.

\subsection{Module 2: 2D Surgical Thread Tracing}
We adapt the analytic cable tracing method from {\textcite{sgtm2}} to trace the path segments from the 2D thread detection masks. However, instead of generating all possible global paths, this work leverages heuristic scoring rules similar to those proposed by 
\textcite{viswanath2023learning} and \textcite{schaal2022} to generate a single global trace. In contrast to the learning-based method proposed in \parencite{viswanath2023learning}, which detects and traces cables simultaneously, we propose an analytical method. 
The method proposed in \parencite{schaal2022} is similar in the sense that it uses scoring functions that prioritize traces which cover more of the cable and have lesser changes in angle. However, they model the thread as a chain of cylinders whereas we fit a 2D spline onto the traced detections to bridge gaps. 
The analytic thread tracer locally traces contiguous segments and greedily stitches them together, as described in Algorithm \ref{alg:trace}.

\begin{algorithm}[]
\caption{2D Surgical Thread Tracing Algorithm}\label{alg:trace}
\begin{algorithmic}
\Require $D \gets \mathrm{pixelwise \ thread \ detection}$
\State $\mathrm{mask} \gets D > \mathrm{thresh_d}$
\State $\mathrm{mask} \gets \mathrm{mask} - (\mathrm{conn \ components \ with \ area < thresh_a}) $
\State $\mathrm{path\_segs} \gets []$
\While {$\mathrm{sum(mask)} > \mathrm{thresh_s}$}
    \State $\mathrm{start\_point} \gets \argmax(\mathrm{D}) $
    \State $\mathrm{paths} \gets \mathrm{sgtm2tracer(mask, start\_point)}$.
    \State $\mathrm{best\_path} \gets \argmax_{\mathrm{p \in paths}}{\mathrm{score(path)}}$
    \State On $\mathrm{mask}$, set points along $\mathrm{best\_path}$ to 0.
    \State Append $\mathrm{best\_path}$ to $\mathrm{path\_segs}$.
\EndWhile

\While {length of $\mathrm{path\_segs} > 1$}
    \State find $i,j$ within $\mathrm{path\_segs}$ with lowest matching cost
    \State $\mathrm{new\_seg} \gets$ merge of $\mathrm{path\_segs[i]}$ and $\mathrm{path\_segs[j]}$
    \State add $\mathrm{new\_seg}$ to $\mathrm{path\_segs}$
\EndWhile

\Return $\mathrm{path\_segs[0]}$
\end{algorithmic}
\end{algorithm}

\subsection{Module 3: 3D Surgical Thread Tracing}
\label{subsec:thread_modeling}
As in prior work~\parencite{thread_reconstruction_cavusoglu}, we model the suturing thread as a 3D NURBS parametric curve. Instead of jointly tracing and reconstructing the thread, we use a dedicated 2D tracer to compute the sequence of thread pixels in both images before reconstructing the 3D thread model. The spline parameter $t \in [0,1]$ describes the normalized distance along the spline. 
To start the 3D tracing method, a 2D NURBS spline defined by 32 control points is fitted to the traces in both images using a least squares approximation. The number of control points is chosen to allow a sufficient amount of flexibility to the spline so that it can approximate tight curves common in suturing thread. 

Next, we triangulate these 2D splines into 3D to estimate the thread state. We therefore propose the following stereo matching approach:
The left trace spline point $p_i^L$ is located at spline parameter $t^L_i$ along the spline and has pixel coordinates $[u^L_i, v^L_i]$ for width and height respectively, starting from the top left corner. For each point along the left spline $p_i^L$, a corresponding point on the right spline $p_{j(i)}^R$ is found which minimizes the difference between spline parameters $t^L_i$ and $t^R_{j(i)}$ and satisfies rectified stereo image properties. Specifically, the right image point should have the same vertical coordinate than the left image point except for a tolerance of up to $\alpha=5$ pixels (condition a). 
 $p_{j(i)}^R$ must be further left within the image than $p_i^L$ (condition b). The right spline candidates must be further along the spline than the last matched right spline point (condition c). $t^L_i$ and $t^R_{j(i)}$ must be within a distance $\beta = 0.05$ (condition d). For a given value of $i$, we seek to solve $$j(i) = \underset{j}{\argmin} |t^R_j-t^L_i|$$ such that a) $|v_i^L - v_j^R| \le \alpha$, b) $u_j^R \le u_i^L$, c) $t^R_{j} > t^R_{j(i-1)} \forall i$, d) $|t^R_j-t^L_i| \le \beta$.

The matched points are then triangulated using the camera intrinsics to obtain their 3D position.
A 3D NURBS spline model is then fitted to the triangulated points using least-squares optimization. 
The values for $\alpha$, $\beta$ and a rejection threshold for bad reconstructions were set empirically as a trade-off between reconstruction quality and number of discarded frames.  

\subsection{Module 4: 3D Surgical Thread Tracking}

Inspired by \textcite{thread_reconstruction_cavusoglu}, we compute 200 correction vectors to update the coordinates of the 3D spline control points between frames. The number of correction vectors was set as a trade-off between tracking accuracy and computation speed. The thread tracking computation time is under 2.5 FPS which is the frame rate of the 2D learned surgical thread detection module as described in Section~\ref{method:detection}. Instead of an energy minimization approach to compute correction vectors, we compute correction vectors using the 2D splines fitted on the current stereo traces. The 2D correction vectors 
are obtained as a sum of two vectors, $c_\text{mask}$ and $c_\text{trace}$. $c_\text{mask}$ is a vector in image space pointing towards the closest point on the prediction mask. 
$c_\text{trace}$ matches the point of the 2D spline fitted on the 2D trace at parameter t with the point at parameter t of the projected 3D spline. 
The 2D correction vectors from both stereo images are triangulated to find 3D correction vectors. Both 3D correction vector terms are averaged to obtain the final set of correction vectors.

Using only the distance correction $c_\text{mask}$, the 3D spline tends to collapse as the segmentation mask of the thread does not constrain the 3D spline along the length of the thread. This is mitigated by the second correction vector, $c_\text{trace}$, which assigns a fully constrained pixel location to each point along the projected 3D spline.

Given the correction vectors, an updated set of control points is computed using the least square control point update described by \textcite{thread_reconstruction_cavusoglu}. 

\subsection{Module 5: Surgical Suture Tail-Shortening}

Initially, the thread passes through the phantom at one suture, with the needle held by one dVRK gripper. Excess thread of between 12-16 cm exists on the needle insertion side of the suture and needs to be pulled through while an unknown amount of thread has already been pulled through the suture. First, the robot uses interactive perception to estimate the needle extraction point by pulling the needle side of the thread taut. This is achieved by moving the needle upwards in positive z direction. The algorithm detects the needle end of the spline to be the one that has the highest z-coordinate. The system detects the taut segment of string by computing the tangent along the 3D spline and identifying a constant tangent direction segment. The angle between the taut segment tangent and the following thread tangent is computed. Pulling the thread upwards leads to a sharp angle in the thread which is identified as the needle extraction point. The 3D thread spline is split into a slack-side and a needle-side at the extraction point and the length of the slack-side thread is computed.

The robot conducts the actual tail-shortening 
by performing a horizontal motion away from the computed wound location (i.e. the extraction point). We continuously run the 3D thread tracking module during this motion, terminating when the thread tail is less than 3cm.
\section{Experiments}
\label{sec:experiments}

\begin{figure}[!t]
    \centering
    \vspace{0.1in}
    \includegraphics[width=0.99\linewidth]{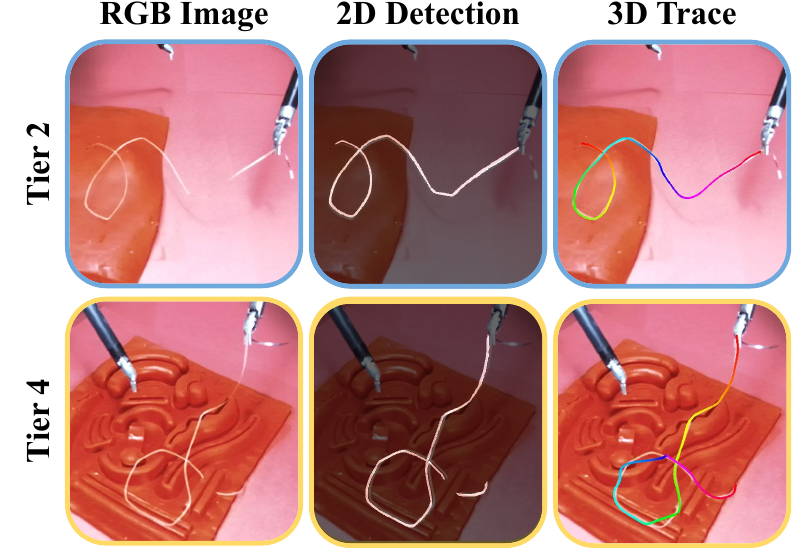}
    \vspace{-0.2in}
    \caption{\textbf{Example 2D Thread Detections and 3D Traces}. 2 example executions of the 3D thread tracing method. \textit{Left} shows the left camera's input RGB image, \textit{Middle} shows the 2D thread detection prediction from the neural network, and \textit{Right} shows the resulting reconstructed 3D spline reprojected into the camera image. The color indicates the path from red to orange.}
    \vspace{-0.15in}
    \label{fig:results}
\end{figure}

\begin{table}[!t]
    \centering
    \newcolumntype{?}{!{\vrule width 1.75pt}}
    \label{tab:exp1}
    \caption{2D Surgical Thread Detection Results}
    \vspace*{-0.1in}
    \setlength\tabcolsep{4pt}
    \begin{tabular}{|c?c|c|c|c|c|} \hline 
    \multicolumn{1}{|c?}{} & \cellcolor{gray!15}{Tier 1} & \cellcolor{gray!15}{Tier 2} & \cellcolor{gray!15}{Tier 3}& \cellcolor{gray!15}{Tier 4} & \cellcolor{gray!15}{Overall}\\ \hline
    \cellcolor{gray!15}{Recall (\%)} & & & & & \\ \hline
    Color thresholding & 32 & 32 & 14 & 40 & 30\\ \hline
    \textbf{Learned segment. (ours)} & \textbf{85} & \textbf{79} & \textbf{80} & \textbf{86} & \textbf{83}\\ \hline
    \cellcolor{gray!15}{Precision (\%)} & & & & & \\ \hline
    Color thresholding & 24 & 27 & 5 & 10 & 17\\ \hline
    \textbf{Learned segment. (ours)} & \textbf{93} & \textbf{93} & \textbf{87} & \textbf{90} & \textbf{91}\\ \hline
    \cellcolor{gray!15}{IoU (\%)} & & & & & \\ \hline
    Color thresholding & 16 & 17 & 4 & 9 & 12\\ \hline
    \textbf{Learned segment. (ours)} & \textbf{80} & \textbf{75} & \textbf{72} & \textbf{79} & \textbf{77}\\ \hline
    \end{tabular}
    \label{tab:detection}
    \vspace*{-0.1in}
\end{table}

\begin{table}[!t]
    \centering
    \newcolumntype{?}{!{\vrule width 1.75pt}}
    \label{tab:exp2}
    \caption{3D Surgical Thread Tracing Results}
    \vspace*{-0.1in}
    \setlength\tabcolsep{4pt}
    \begin{tabular}{|c?c|c|c|c|c|} \hline 
    \multicolumn{1}{|c?}{} & \cellcolor{gray!15}{Tier 1} & \cellcolor{gray!15}{Tier 2} & \cellcolor{gray!15}{Tier 3}& \cellcolor{gray!15}{Tier 4} & \cellcolor{gray!15}{Overall}\\ \hline
    Mean Reproj. Err. (pix) & 0.58 &   1.14 & 2.50 & 1.10 &1.33\\ \hline
    Max Reproj. Err. (pix) & 13.34 &   20.61 & 62.16 & 43.17 &62.16\\ \hline
    \end{tabular}
    \label{tab:reconstruction}
    \vspace*{-0.2in}
\end{table}

\subsection{Modules 1-3: 2D Thread Detection and 2D  \&  3D Tracing}
\label{sec:collection}
\subsubsection{Setup}
We test the first 3 modules by using the workspace described in Sec~\ref{sec:workspace_setup} 
The experiments begin with the needle in the right end-effector and the tip of the thread going through the phantom. We collect test examples from 4 difficulty tiers: \\
\textbf{Tier~1}: No self-intersection in thread, reversed phantom. \textbf{Tier~2}: $\ge1$ self-intersection in thread, reversed phantom. \textbf{Tier~3}: No self-intersection in thread, phantom facing up. \textbf{Tier~4}: $\ge1$ self-intersection in thread, phantom facing up.
We collect and label stereo images for 5 scenes per tier for a total of 10 images per tier.

\subsubsection{Metrics}
\label{subsubsec:collection_metrics}
In this experiment, we evaluate the IoU, precision, and recall metrics of the segmentation mask with respect to human-labeled ground truth segmentation masks. To evaluate the 3D model of the thread,
we report the reprojection error between the human-labeled ground truth thread segmentations and the projection of the 3D model of the thread into both stereo images. 

\subsubsection{Results and Failure Modes}
Results for 2D thread detection and 3D thread tracing are presented in Table~\ref{tab:detection} and Table~\ref{tab:reconstruction} respectively. Example detection masks and reconstructions are shown in Figure~\ref{fig:results}. Comparison with the color thresholding baseline clearly shows that the learned method is able to detect thread in low contrast scenes and in the presence of light reflections on the phantom. Note that while detections can miss segments of highly difficult thread (recall of 83\%), the precision of predictions is $90\%$.
Our detection model shows slightly better recall and IoU performance in the more difficult Tier 4 with respect to Tier 2. This difference is due to a scene in Tier 2 in which the thread has a particularly low contrast with the background and is thus not detected. The discrepancy lies in the error bars of this experiment. \textcite{thread_reconstruction_liu} report an average IoU of 85\% with a recall of 93\%. While these are impressive results, 
all their scenes have suturing threads lying on the ground plane without any tools that can cast shadows or reflective configurations that make segmentation more challenging. 

The 3D surgical thread traces have an overall mean error of 1.33 pixels, showing that the reconstruction approximates the spline well in general. The 3D tracing fails on parts of the thread in 3 scenes, resulting in the high maximum reprojection error. These errors are mainly due to reflective bright edges on the phantom which lead to erroneous detections and 3D traces. 
\textcite{thread_reconstruction_yip}, report reprojection errors between mean 0.4 and 1.1 pixels on 10 real scenes with printed surgical backgrounds. These results seem comparable to ours even though performance remains highly dependant on the particular scenes, making results hard to compare objectively.

\subsection{Module 4: 3D Surgical Thread Tracking}
\label{sec:exp-module-4}
\subsubsection{Setup}

Using the workspace setup described in Section \ref{sec:workspace_setup}, we thread the needle through the phantom and place it in the right gripper of the dVRK. We evaluate the thread tracking system on two trajectories: the ``no loop'' trajectory, a line in the image plane, and the ``one loop'', an elliptical track above the phantom. The first trajectory avoids any occlusion or self-intersection of the thread, while the second incurs a challenging, self-crossing configuration.

\subsubsection{Metrics}
We report the same metrics as in the single-frame 3D tracing (Section~\ref{subsubsec:collection_metrics}) experiments. 
We manually label ground-truth segmentation masks in stereo images taken every 1~cm along the trajectory and evaluate the 3D thread model against them. This leads to 10 evaluation frames for the ``No loop'' trajectory and 9 for the ``One loop.''

\subsubsection{Results and Failure Modes}
The results in Table~\ref{tab:tracking} suggest that the method is able to track the thread reliably in both configurations. 
The ``One loop'' trajectory sees higher reprojection errors, as it presents a more challenging thread configuration for the tracer. 
The full tracking pipeline achieves a mean reprojection error of 0.39 pixels on the intersection-free trajectory and a 1.28 pixels on the loop-forming trajectory. \textcite{thread_reconstruction_hager_1}, report a mean reprojection error of 1.21 pixels. However, they use only short threads in configurations which present no self-intersections. 

\begin{table}[!t]
    \vspace{0.1in}
    \begin{center}
        \newcolumntype{?}{!{\vrule width 1.75pt}}
        \label{tab:exp3}
        \caption{3D Surgical Thread Tracking Results}
        \setlength\tabcolsep{4pt}
        \begin{tabular}{|c?c|c|} \hline 
        \multicolumn{1}{|c?}{ } & \cellcolor{gray!15}{Mean Reproj. Err.} & \cellcolor{gray!15}{Max Reproj. Err.} \\ \hline
        \cellcolor{gray!15}{No loop} & & \\ \hline
        No tracking & \textbf{0.30 pix} & 9.22 pix \\ \hline
        With tracking & 0.39 pix & \textbf{5.39 pix} \\ \hline
        \cellcolor{gray!15}{One loop} & & \\ \hline
        No tracking & 5.37 pix & 94.92 pix \\ \hline
        With tracking & \textbf{1.28 pix} & \textbf{16.26 pix} \\\hline
        \end{tabular}
        \label{tab:tracking}
    \end{center}
    \textit{No tracking} refers to computing a new 3D trace for every frame. \\ \textit{With tracking} refers to using Module 4 and leads to a significantly better mean reprojection error in the more difficult ``One loop'' case.
    \vspace*{-0.25in}
\end{table}

\subsection{Module 5: Surgical Suture Tail-Shortening}
\subsubsection{Setup}
We additionally test automated suture tail shortening using the workspace setup described in Section \ref{sec:workspace_setup}, with the needle threaded fully through the reversed phantom and held by the right end-effector. The tail of the thread is then placed arbitrarily in the workspace such that the entire suture thread is within the view of both the right and left stereo cameras.

\subsubsection{Metrics}
We define a successful tail-shortening maneuver as a termination in which the final tail length lies within 1~cm of the desired value of 3~cm. We report the success rate of our pipeline on the tail-shortening task, as well as the mean absolute error between the achieved and desired length and the average time to completion for this task.

\subsubsection{Results and Failure Modes}
The proposed method achieves 18 successes out of 20 trials with a mean absolute tail error of 0.53cm. The mean time to completion is 106.8 seconds. The method achieves a success rate of 90\%, indicating that our pipeline provides high-confidence 3D traces using interactive perception. The main failure case occurs during the 3D tracing of the spline due to false detections on the human organ phantom. 

\section{Limitations \& future work}
The primary limitation is the uncertainty about the ability of the learned 2D surgical thread detection to generalize to new thread or phantoms, which will be addressed in future work.
Also, the learned thread segmentation method remains vulnerable to false positive detections of light reflections from the edges of the human organ phantom. This could be mitigated in future work by adapting the lighting setup of the workspace.

\footnotesize{
    \section*{Acknowledgement}
This research was performed at the AUTOLAB at UC Berkeley in affiliation with the Berkeley AI Research (BAIR) Lab, and the CITRIS "People and Robots" (CPAR) Initiative. The da Vinci Research Kit is supported by the National Science Foundation, via the National Robotics Initiative (NRI), as part of the collaborative research project "Software Framework for Research in Semi-Autonomous Teleoperation" between The Johns Hopkins University (IIS1637789), Worcester Polytechnic Institute (IIS 1637759), and the University of Washington (IIS 1637444).)
}

\printbibliography

\clearpage
\normalsize

\section{Appendix}
\label{sec:appendix}

\subsection{Workspace}
\begin{figure}[h]
    \centering
    \includegraphics[width=0.9\linewidth]{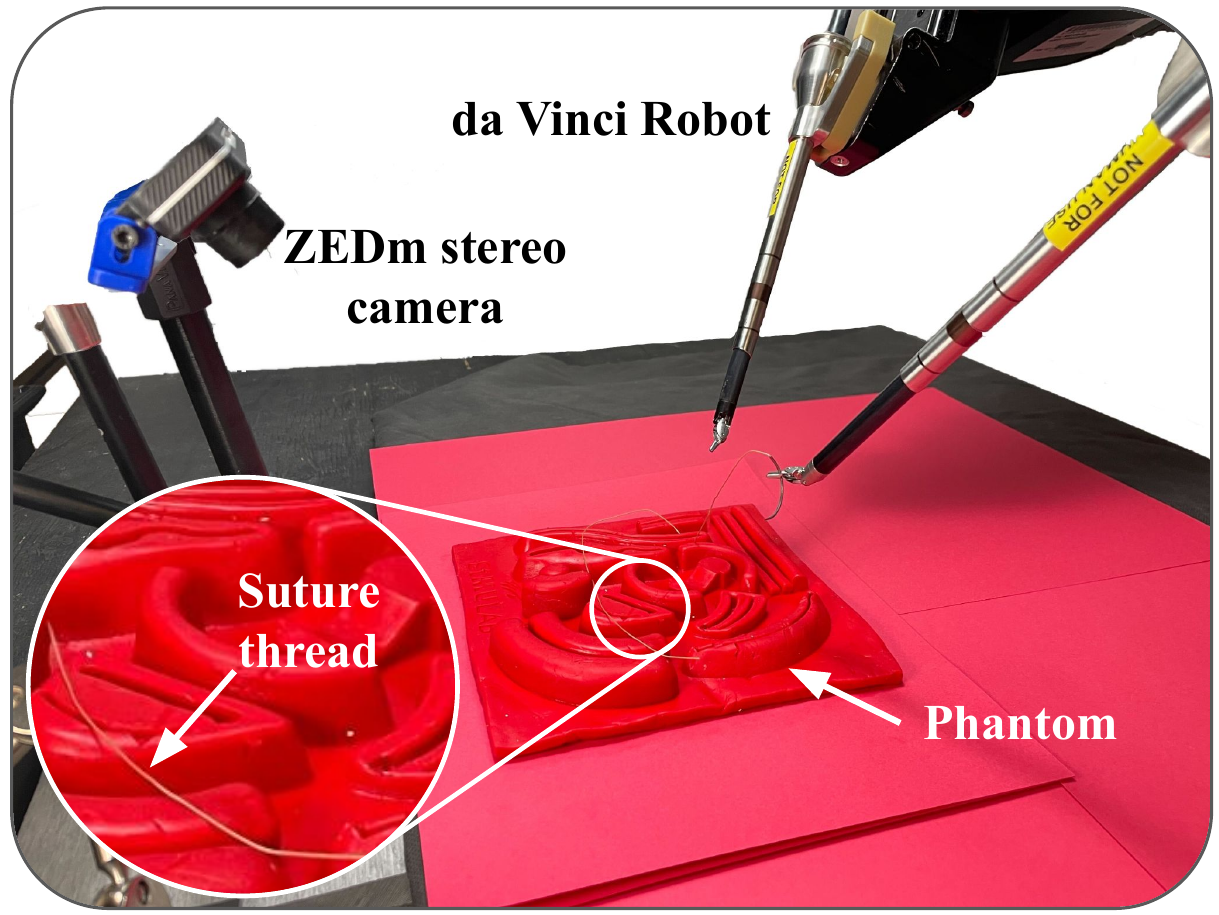}
    \vspace*{-0.1in}
    \caption{\textbf{Workspace}. The da Vinci Research Kit with two PSMs (Patient Side Manipulators) holds a thread above the phantom tissue placed on a red background. The suture thread is thin, making it difficult to perceive. The ZEDm stereo camera pair faces the workspace on the opposite side to the robot at a steep downwards angle.}
    \label{fig:workspace}
    \vspace*{-0.2in}
\end{figure}

Figure~\ref{fig:workspace}, shows an image of the workspace used for the experiments. It contains two dVRK robot arms, a ZEDm stereo camera, a red tissue phantom placed on a red background and a surgical suture composed of a needle and white thread.

\subsection{Module 4 Experiment Extension: 3D Surgical Thread Tracking}

We show an ablation study for the experiment described in Module 3 \ref{sec:exp-module-4}. It shows the value of both components of the 3D surgical thread tracking correction vector.

\subsubsection{Setup}
\begin{figure}[h!]
    \centering
    \includegraphics[width=0.99\linewidth]{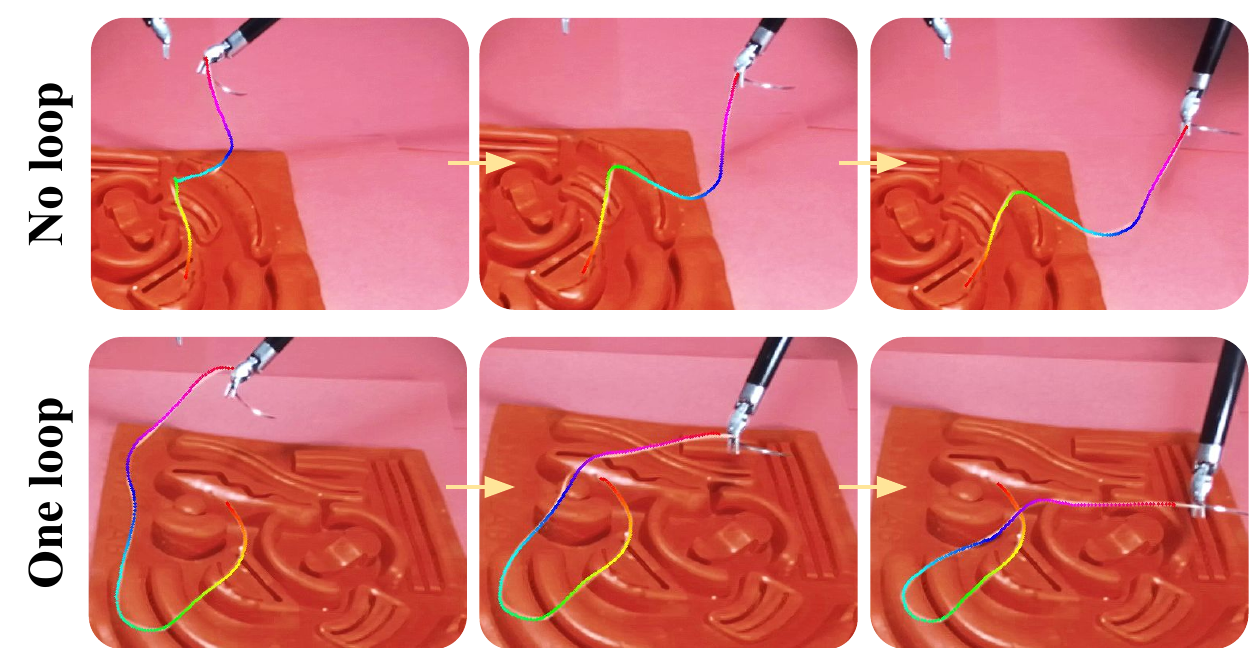}
    \vspace{-0.2in}
    \caption{\textbf{Tracking Experiment Trajectories}: these panels illustrate the two trajectories used for computing tracking performance. The rainbow colored line shows the projected 3D spline while the colors depicts the sequence of the spline.\textit{Top} shows the first "No loop" trajectory where the thread has no self-intersections (moving from left to right). \textit{Bottom} shows the second "One loop" trajectory where a self-intersecting thread configuration is created.}
    \vspace{-0.2in}
    \label{fig:tracking}
\end{figure}

The setup is the same as in the main experiment described in Section \ref{sec:exp-module-4}. We evaluate the performance on the same trajectories we call `No Loop' and `One Loop'. Both trajectories are represented in Figure \ref{fig:tracking}.

\subsubsection{Metrics}
We use the same metrics as in the main experiment, i.e. mean and maximum reprojection error in pixels. Also, we add the ground truth coverage metric which measure how much of the ground truth segmentation mask is covered by the reprojected 3D surgical thread model spline. 

\subsubsection{Results and Failure Modes}
\begin{figure}[h!]
    \centering
    \includegraphics[width=0.99\linewidth]{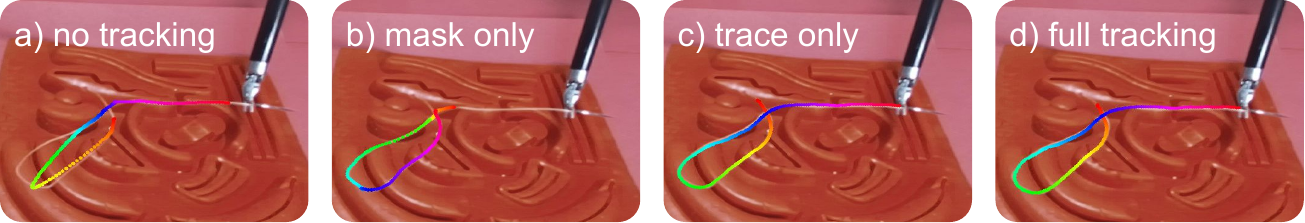}
    \vspace{-0.2in}
    \caption{\textbf{Tracking Ablation Study}: This figure shows a scene from the \textit{One loop} trajectory with the 3D spline reprojected into the left camera image computed with four tracking ablations for which experimental results are in Table~\ref{tab:tracking_ablation}. Image a shows the spline obtained through reconstruction only, image b shows the spline obtained when only the mask error $c_\text{mask}$ is used for tracking. Image c shows the spline obtained when only the trace error $c_\text{trace}$ is used for tracking and image d shows the result obtained with the full tracking method. It can be seen in image b that the spline collapses along the detection mask when the trace correction vector is not used as well.}
    \vspace{-0.1in}
    \label{fig:tracking_ablation}
\end{figure}

We present three ablations of the 3D surgical thread tracking pipeline and compare them to the full pipeline performance. Results can be seen in Table \ref{tab:tracking_ablation}. \textit{No tracking} refers to performing a new 3D reconstruction every frame. It can be seen that especially in the more difficult `One Loop' trajectory, not using the knowledge from prior frames leads to higher reprojection errors. The \textit{Mask track} ablation refers to tracking with only the correction vector $c_\text{mask}$. While it leads to lower reprojection errors, looking at the median ground truth coverage shows that the spline actually shortens along the detection mask as the spline points are not fixed along this degree of freedom. This is clearly visible when looking at Figure \ref{fig:tracking_ablation}. The \textit{Trace track} ablation refers to tracking with only the correction vector $c_\text{trace}$. This correction vector addresses the spline collapse which arises when only using $c_\text{mask}$ as it fully constrains the 3D position of the spline evaluation points. However, it can be seen that using only this term leads to higher reprojection errors as the two traces on the left and right image can be of different length due to perspective effects, thus leading to a mismatch of correction vectors between both images. This imprecision in turn is addressed by the first correction vector.
Using the \textit{Full tracking} method leads to the best mean reprojection error while achieving a high ground truth coverage value which shows that it is reconstructing the whole thread.

\begin{table}[h!]
    \vspace*{-0.1in}
    \begin{center}
        \newcolumntype{?}{!{\vrule width 1.75pt}}
        \caption{3D Surgical Thread Tracking Results}
        \setlength\tabcolsep{4pt}
        \begin{tabular}{|c?c|c|c|} \hline 
        \multicolumn{1}{|c?}{ } & \cellcolor{gray!15}{Mean Reproj. Err.} & \cellcolor{gray!15}{Max Reproj. Err.} & \cellcolor{gray!15}{GT Coverage} \\ \hline
        \cellcolor{gray!15}{No loop} & & & \\ \hline
        No tracking & 0.30 pix & 9.22 pix & 93.22 \% \\ \hline
        Mask track. & \textbf{0.20 pix} & \textbf{4.12 pix} & 88.56 \% \\ \hline
        Trace track. & 0.77 pix & 7.00 pix & \textbf{96.37 \%}  \\ \hline
        Full tracking & 0.39 pix & 5.39 pix & 96.10 \% \\ \hline
        \cellcolor{gray!15}{One loop} & & & \\ \hline
        No tracking & 5.37 pix & 94.92 pix & \textbf{95.90 \%}  \\ \hline
        Mask track. & \textbf{0.15 pix} & \textbf{5.0 pix} & 82.79 \% \\ \hline
        Trace track. & 1.91 pix & 17.03 pix & 94.53 \%  \\ \hline
        Full tracking & 1.28 pix & 16.26 pix & 95.26 \% \\\hline
        \end{tabular}
        \label{tab:tracking_ablation}
    \end{center}
    \textit{GT Coverage} is the ground truth coverage metric which measures how much of the ground truth mask overlaps with the current reprojected thread model in pixel space. \textit{No tracking} refers to performing a new 3D reconstruction for every frame. The \textit{Mask track.} ablation refers to tracking with only the correction vector $c_\text{mask}$. The \textit{Trace track.} ablation refers to tracking with only the correction vector $c_\text{trace}$.  Using the \textit{Full tracking} method leads to the best mean reprojection error while reconstructing the whole thread.
    \vspace*{-0.25in}
\end{table}

\end{document}